\documentclass[conference,]{IEEEtran}
\IEEEoverridecommandlockouts
\usepackage[utf8]{inputenc}
\usepackage[numbers, sort]{natbib}
\usepackage{amssymb}
\usepackage{MnSymbol}
\usepackage{accents}
\usepackage{verbatim}
\usepackage{listings}
\usepackage[inline]{enumitem}

\usepackage{amsmath}
\DeclareMathOperator*{\argmin}{arg\,min}
\usepackage[dvipsnames]{xcolor}
\usepackage{soul}

\usepackage{booktabs}
\usepackage{multirow}
\usepackage{caption}
\usepackage{subcaption}
\usepackage{floatrow}
\usepackage{tikz}
\usetikzlibrary{shapes, shapes.misc, shapes.geometric, arrows, calc, positioning, fit}
\title{
	Interoperability and machine-to-machine translation model with mappings to machine learning tasks
	\thanks{This work is partly supported by the H2020 research and innovation programme ECSEL Joint Undertaking and Vinnova, project no. 737459, and by The Kempe Foundations under contract SMK-1429.
	}
}
\author{\IEEEauthorblockN{Jacob Nilsson, Fredrik Sandin and Jerker Delsing}
	\IEEEauthorblockA{EISLAB, Lule{\aa} University of Technology\\ 971 87 Lule{\aa}, Sweden}
}
\newcommand{\ifour}{Industry~4.0}
	\date{January 2019}

\begin{document}

	\maketitle
	\begin{abstract}
		Modern large-scale automation systems integrate thousands to hundreds of thousands of physical sensors and actuators.
		Demands for more flexible reconfiguration of production systems and optimization across different information models, standards and legacy systems challenge current system interoperability concepts.
		Automatic semantic translation across information models and standards is an increasingly important problem that needs to be addressed to fulfill these demands in a cost-efficient manner under constraints of human capacity and resources in relation to timing requirements and system complexity.
		Here we define a translator-based operational interoperability model for interacting cyber-physical systems in mathematical terms, which includes system identification and ontology-based translation as special cases.
		We present alternative mathematical definitions of the translator learning task and mappings to similar machine learning tasks and solutions based on
		recent developments in machine learning. %
		Possibilities to learn translators between artefacts without a common physical context, for example in simulations of digital twins and across layers of 
		the automation pyramid are briefly discussed.
	\end{abstract}

\section{Introduction}

Automation systems in {\ifour} \cite{lasi2014industry4,hankel2015reference} and the Internet-of-Things (IoT) \cite{borgia2014internet} are designed as networks of interacting elements, which can include thousands to hundreds of thousands of physical sensors and actuators.
Efficient operation and flexible production require that physical and software components are well integrated, and increasingly that such complex automation systems can be swiftly reconfigured and optimized on demand using models, simulations and data analytics \cite{ierc2013iot,wang2016smarti4}.
Achieving this goal is a nontrivial task, because it requires interoperability of physical devices, software, simulation tools, data analytics tools and legacy systems from different vendors and across standards
\cite{nilsson2018semantic,gurdur2018systematic,derhamy2017iot,ierc2013iot}.
Standardisation of machine-to-machine (M2M) communication, like the OPC Unified Architecture  (OPC~UA\footnote{https://opcfoundation.org/about/opc-technologies/opc-ua/}) \cite{leitner2006opc} which offers scalable and secure communication over the automation pyramid, and development of Service Oriented Architectures (SOA), like the Arrowhead Framework \cite{arrowhead2019}, are developments supporting the vision of interoperability in {\ifour} and the IoT.

However, in addition to data exchange by protocol-level standardisation and translation \cite{derhamy2017iot}, information models are required to correctly interpret and make use of the data.
There are many different information models and standards defining semantics of data and services, which are developed and customized to fit different industry segments, products, components and vendors.
This implies that the problem to translate data representations between different domains is increasingly relevant for robust on-demand interoperability in large-scale automation systems.
This capacity is referred to as {\it dynamic interoperability} \cite{ierc2013iot}, and {\it operational interoperability} \cite{gurdur2018systematic} meaning that systems 
are capable to access services from other systems and use the services to operate effectively together.
Thus, focus needs to shift from computing and reasoning in accordance with a representational system to automatic translation and computing over multiple representational systems \cite{licato2017representation,ierc2013iot} and engineers should 
operate at the levels where system-of-systems goals and constraints are defined.

In this paper, we outline a mathematical model of the problem to translate between representational systems in cyber-physical systems (CPS) with integrated physical and software components, and we map some alternative definitions of the translation problem in this model to machine learning tasks and the corresponding state-of-the-art methods.
In this model, concepts like symbol grounding, semantics, translation and interpretation are mathematically formulated and possibilities to more automatically create semantic translators with machine learning methods are outlined.
\section{Interoperability model}

When integrating SOA systems and services, which are designed according to different standards and specifications, various interfaces that are also subject to domain-specific assumptions and implementation characteristics need to be interconnected.
It is common practice to engineer the connections between such interfaces in the form of software {\it adapters} that make different components, data, services and systems semantically interoperable, so that functional and non-functional system requirements can be met.
This way, a modular structure is maintained, which makes testing and the eventual replacement of a module and updates of the related adapters tractable in otherwise complex systems, at the cost of a quadratic relationship between the number of adapters and the number of interfaces.

In deterministic protocol translation, where representational and computational completeness allows for the use of an intermediate ``pivot’’ representation of information, the quadratic complexity of the adapter concept can be reduced to linearity, see for example \cite{derhamy2017iot} and \cite{malo2013hub}.
However, in the case of semantic translation considered here, it is not clear that such universal intermediate representations exist and constitute a resource-efficient and feasible approach to translation.
Furthermore, the research field of dynamic and operational interoperability in SOA lacks a precise mathematical formulation and consensus about the key problem(s).
Therefore, we approach the translation problem by formulating it in precise mathematical terms that can be mapped to machine learning tasks.

We define the M2M interoperability problem in terms of {\it translator} functions, $T^{AB}$, which map messages, $m^A$, from one domain named CPS~A to messages in another domain, $m^B$, named CPS~B, see Figure~\ref{fig:interop}.
The translators can be arbitrarily complex functions that are generated as integrated parts of the overall SOA, thereby maintaining a modular architecture as in the case of engineered adapters.
In general, the translated messages, $\hat{m}^B$, cannot be semantically and otherwise identical to the messages communicated within CPS~B, $m^B$, but we can optimize the translator functions to make the error small with respect to an operational loss or utility function.
In the following, we elaborate on the latter point and introduce the additional symbols and relationships of the model as the basis for defining translator learning tasks, which in principle can be addressed with machine learning methods.

\tikzset{system/.style = {rectangle,
		minimum height = 0.25\columnwidth,
		align=center,
		minimum width = 0.24\columnwidth,
		text width = 0.18\columnwidth,
		rounded corners
	}
}
\pgfdeclarelayer{background}
\pgfdeclarelayer{foreground}
\pgfsetlayers{background,main,foreground}
\newcommand{\arrowoffsetx}{0.12\columnwidth}
\newcommand{\arrowoffsety}{0.031\columnwidth}
\begin{figure}[t]
	\centering
	\begin{tikzpicture}[>=latex]%
		\node[system] (A) {};
		\node[below right, align=center, text width = 0.22\columnwidth] at (A.north west)
			{CPS A\\[0.2em]
				$\mathrm{x}^A,
				G^A,
				\mathrm{m}^A$\\[0.07\columnwidth]
				$u^A, y^A$
			};
		\node[system, right = 0.35\columnwidth of A] (B) {};
		\node[below right, align=center, text width= 0.22\columnwidth] at (B.north west)
			{CPS B\\[0.2em]
				$\mathrm{x}^B,
				G^B,
				\mathrm{m}^B$\\[0.07\columnwidth]
				$u^B, y^B$
			};		
		\coordinate (physical offset) at (0, -0.06\columnwidth);
		\coordinate (B separation) at ($(B.west)+(physical offset)$);
		\coordinate (B separation east) at ($(B.east)+(physical offset)$);
		\coordinate (A separation) at ($(A.west)+(physical offset)$);
		\path (A) to node[draw, above = 0.035\columnwidth] (translator) {$T^{AB}$} (B);
		\path[->, draw, above] (A.east |- translator) to node (mAB-1) {$\mathrm{m}^A$} (translator.west);
		\path[->, draw, above] (translator.east) to node (mAB-2) {$\hat{\mathrm{m}}^{B}$} (translator.east -| B.west);
		\path (A separation) to node[] (physical) {$u, y$} (B.south east);
		\draw[<->, rounded corners] (B.west |- physical)+(\arrowoffsetx, \arrowoffsety) |- (physical.east);
		\draw[<->, rounded corners] (A.east |- physical)+(-\arrowoffsetx,\arrowoffsety) |- (physical.west);
		\begin{pgfonlayer}{foreground}
			\draw[rounded corners] (A.west) |- (A.north east) |- (B separation) |- (B.north east) |- (A.south west) -- (A.west);
		\end{pgfonlayer}
		\begin{pgfonlayer}{background}
			\coordinate (semiphys offset) at (0, 0.01\columnwidth);
			\path[fill=black!5] (B.west) |- ($(B.east)+(semiphys offset)$) |- (B.east)   |- (A separation) {[rounded corners] -| (B.west)};
			\path[fill=black!5] (A.west) |- ($(A.east)+(semiphys offset)$) |- (A.east) {[rounded corners] |- (B separation)} -- (A separation);
			\path[fill=black!10] (A separation) -- (B separation east) {[rounded corners] |- (A.south) -| (A separation)};
			\path[draw, gray!60] ($(B.west)+(semiphys offset)$) -- ($(B.east)+(semiphys offset)$);
			\path[draw, gray!60] ($(A.west)+(semiphys offset)$) -- ($(A.east)+(semiphys offset)$);
			\path[draw, gray!80] (A separation) -- (B separation east);
		\end{pgfonlayer}
	\end{tikzpicture}
	\caption{Model of communicating cyber-physical systems (CPS) with different data representations and semantic definitions that interact in a physical environment (gray) and service-oriented architecture (white) via messages $m$ translated by a function $T^{AB}$.
	}
	\label{fig:interop}
\end{figure}
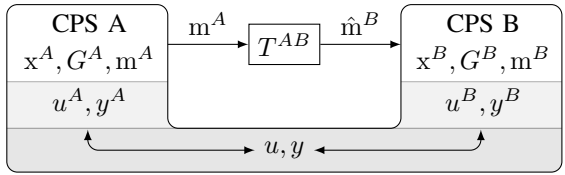
 
The model is divided in three levels: cyber (white), physical representation (light gray) and the shared physical environment (gray), see Figure~\ref{fig:interop}.
At the cyber level, the graphs $G^A$ and $G^B$ define all discrete symbolic and sub-symbolic metadata that is specific for CPS~A and CPS~B, respectively.
For example, the nodes and edges of these graphs can represent subject, predicate, and object semantic triples defined in the Resource Description Framework (RDF).
Each CPS also has discrete internal states, $x^A$ and $x^B$ respectively, such as the computer program variables of all devices in a CPS, which are not directly readable or writeable in the SOA but may be read and modified indirectly via the messages and services.
The environment has inputs, $u$, which can be affected by actuator devices, and outputs, $y$, which can be measured with sensor devices.
In CPS~A, the outputs of the sensor devices are represented at the cyber level as discrete variables $y^A$ and the actuators are controlled by discrete variables $u^A$, and similarly for CPS~B.
From the viewpoint of causality, $u$ influences $y$ and thus changes of elements of $u^A$ may influence the values of elements in both $y^A$ and $y^B$, and vice versa.

Messages are generated by encoder functions on the form
\begin{equation}
	m^A \leftarrow E^A(u^A, y^A, x^A; G^A),
	\label{eq:encoder}
\end{equation}
which typically are implemented in the form of computer programs.
Similarly, the internal states are updated by decoder functions
\begin{equation}
	(x^A,u^A) \leftarrow D^A(m^A; x^A, u^A, y^A; G^A),
	\label{eq:decoder}
\end{equation}
which are matched to the corresponding encoder functions.
However, a decoder $D^B$ can in general not be combined with an encoder $E^A$, and vice versa.

Although some technical details and challenges are hidden in this abstract model, the model enables us to define concepts and relationships that otherwise are ambiguous and described differently in the literature depending on the context.
The task to model dynamic relationships between $u$ and $y$ in terms of $u^A$ and $y^A$ (or $u^B$ and $y^B$ etc) is the central problem of system identification \cite{ljung2010perspectives}. 
The task to model and control one CPS in terms of the relationships between $u^A$, $y^A$, $x^A$ and sometimes also $G^A$ is more complex \cite{graja2018comprehensive} and typically involves hybrid models with state-dependent dynamic descriptions.
This is a central problem in automatic control and CPS engineering.

Symbol grounding \cite{cubek2015grounding} refers to the relations between a symbol defined by $G^A$ and the related discrete values of $\{x^A,u^A,y^A\}$ (similarly for $G_B$) and the property of the environment $\{u,y\}$ that the symbol represents.
A {\it grounding problem} appears when a symbol defined in $G^A$ have an underfitted relationship to the referenced property of the environment represented via $\{x^A,u^A,y^A\}$ (similarly for $G_B$), such that symbols in $G^A$ and $G^B$ cannot be conclusively compared for similarity although both systems are defined in the same environment.
Therefore, symbol grounding is just as relevant for translator learning as it is for reliable inference in cognitive science and artificial intelligence.

Listing~\ref{lst:data} presents two examples of SenML messages that are constructed to illustrate the character of a semantic translation problem,
$\hat{m}^B = T^{AB}(m^A)$.
\begin{lstlisting}[float=b, caption={Two semantically similar but machine-incompatible messages. Parts with the same color describe the same concept, property or object.}, label={lst:data}, xleftmargin=\parindent, gobble=2, tabsize=2, basicstyle=\scriptsize\bfseries\ttfamily, escapechar=@]
	# System A message
	[
		{"bn":"@\color{green}127.0.0.1/temp-service@","bt":1549359472},
		{"u":"lon","v":@\color{red}65.61721@},
		{"u":"lat","v":@\color{red}22.13683@},
		{"u":"K","v",@\color{SkyBlue}253@}
	]

	# System B message
	[
		{"n":"@\color{red}office-A2312@-@\color{green}temp-sensor@",
		 "u":"Cel",
		 "v":@\color{SkyBlue}-20.4@,
		 "t":1549359472}
	]
\end{lstlisting}
Both messages encode information about the temperature in one office at our university and thus represents related physical properties.
A and B can for example refer to the heating and ventilation systems in the office, respectively, and thus the temperatures are not necessarily identical.
The message from System~A includes the service URI and the time, longitude and latitude of the temperature measurement with unit `K' for Kelvin and numeric value $293$.
The message from System~B includes the name of the temperature sensor, the unit ``Cel'' for Celsius, the value $-20.4$ and the time of the temperature measurement.

A translator, $T^{AB}$, could in this scenario for example be used by an indoor climate and energy optimization service that is capable to interpret messages of the second kind in Listing~\ref{lst:data}, but not of the first kind.
By using the translator this service could for example improve the quality of the indoor climate, further reduce the energy used, or be more fault resilient in case of a sensor fault.
As outlined above, messages encoded by CPS~A in Figure~\ref{fig:interop} can in general not be correctly interpreted by CPS~B, and vice versa.
How can a translator that solves this problem be automatically generated?

We approach this problem by defining a computable function, $J$, that determines to what extent the system-of-systems (SoS) formed by CPS~A, CPS~B etc fulfils particular operational requirements and goals.
For example, the function $J$ could be formulated as a loss function in machine learning, or a utility function of a multi-agent system,
and the translator learning task is to minimize the loss or maximize the expected utility.
Some possible definitions of the function $J$ are listed in Table~\ref{lossfunctions}.
\begin{table}
    \footnotesize
	\caption{Examples of loss/utility functions.}
	\centering
	\begin{tabular}{lll}
		\toprule
		Type & Function & Example \\
		\midrule
        Causation & $J(y^A, u^B)$, $J(u^A, y^B)$ & Step responses \\
        Correlation & $J(y^A, y^B)$ & Related measurements \\
        Abstract & $J(x^A, x^B)$ & Efficiency optimization \\
		\bottomrule
	\end{tabular}
	\label{lossfunctions}
\end{table}
The key points are that engineering resources are focused on defining $J$ in terms of SoS goals and requirements, and that it is possible to optimize $J$ by defining and updating $T^{AB}$ using machine learning methods on the form
\begin{align}
	\hat{m}^B &= T^{AB}(m^A; G^A, G^B;\ldots), 
	\label{eq:translator} \\
    T^{AB} &= \argmin_{T^{AB}}~J\left(y^A, u^B\!(m^B;\ldots;\hat{m}^{B})\right),
    \label{eq:loss}
\end{align}
and similarly for other choices of $J$ and in the case of expected utility maximization.

For example, in the office example introduced above,
$J$ could be a causality type loss and $T^{AB}$ could be a recurrent neural network, which is trained until the ventilation system decodes $\hat{m}^B$ so that the effects of varying $u^B(m^B;\ldots;\hat{m}^{B})$ on $y^A$ are correctly predicted across instrumented offices.

In general, the translator function should depend on symbols in $G^A$ and $G^B$, and it can depend also on other information sources, like public datasets \cite{kolyvakis2018deepalignment} and historical CPS data used to fit sub-symbolic relationships more accurately.
In principle, the translator $T^{AB}$ can be considered to perform three tasks:
\begin{enumerate} 
	\item Estimate the decoder, $D^A$.
	\item Map information from domain A to domain B. %
	\item Estimate the encoder, $E^B$.
\end{enumerate}
Like in the field of machine translation of natural language we can attempt to explicitly model these individual mappings, or we can model the overall mapping $T^{AB}$.
We elaborate on machine learning tasks and methods that may be useful to address the translator learning task outlined above after briefly introducing some related work in the next section.

\section{Related work on interoperability solutions}

To fully exploit the potential of the IoT and Industry4.0, engineering resources should to a larger extent be focused on high-level benefits of interoperability and system integration \cite{ierc2013iot}.
Automated approaches to establish and maintain interoperability are needed to enable on-demand service composition and meet the demands for flexible production and high efficiency given the high complexity and diversity of automation systems driven by the rapid technological development \cite{nilsson2018semantic}.
Architectures similar to the model presented here have been independently developed by \citet{malo2013hub} who describe architectures that allow for maximum interoperability.
Our model describes the specific task to translate between services and data formats, but can in general be considered as a special case of the architectures considered in that work.
Concepts and methods developed for the semantic web \cite{shadbolt2006semantic} are widely used to integrate human- and machine-readable metadata to support the adapter engineering and system integration processes, such as ontologies, ontology alignment and ontology-based reasoning engines.
The semantic web tags websites with ontological metadata, typically encoded in RDF or higher-level ontology languages like the Web Ontology Language (OWL).
The Semantic Sensor Network (SSN) \cite{compton2012ssn}, an ontology specialized for describing sensors, is one example of a domain-specific ontology.
The Open Semantic Framework (OSF) \cite{mayer2019open} combines many such specific ontologies into an extendable framework, fusing both general and domain specific knowledge.
Ontologies form the core of semantic technologies, but not all ontologies can be combined and function together.
Ontologies that are based on different standards and definitions can model related physical and cyber entities in different ways, thus leading to contradictions and under-determined relationships between symbols when different technologies are combined.

In addition to {\it semantic interoperability}, which focuses on supporting the engineering process with such standardized metadata models, methods and tools for automatic on-demand dynamic interoperability \cite{ierc2013iot} and operational interoperability \cite{gurdur2018systematic} are developed. 
Symbolic reasoners can be applied to create Web-like mashups in highly dynamic environments \cite{kovatsch2015practical}, but suffers from state-space explosion when physical states are included.
This challenge is recognized also in the domain of symbolic artificial intelligence. 
Furthermore, automatic reasoning in terms of symbolic metadata is unreliable in complex and uncertain real-world environments because symbolic data does not include all necessary information about the context, environment and system (cf. comments on symbol grounding and underfitted symbol relationships in the former section).
Therefore, ontology-based translation is extended with sub-symbolic mapping and reasoning mechanisms.
A recent example in this direction is deep alignment of ontologies \cite{kolyvakis2018deepalignment}.
Deep alignment enables discovery of sub-symbolic mappings between elements of ontologies by a data-driven optimization method,
where textual descriptions are represented by word-vectors learned from an auxiliary data set, similar to techniques used in natural language processing.

The development of more potent interoperability methods and technologies are of central importance for modern SOA, like the aforementioned Arrowhead Framework.
For example, ontology-based XML-message translation has been extended with semantic annotations \cite{moutinho2018extended}, see also former work in \cite{malo2013hub}.
That translator can map elements, perform unit conversion, detect missing data and, in certain cases, find and add the missing data.
Another example is the architecture for device management using autonomic computing \cite{lam2018supporting}, where a manager monitors and plans execution using ontologies and a reasoning engine.

Data lakes, like the The Big Data Europe platform\footnote{https://www.big-data-europe.eu/platform/}, is another approach where heterogeneous data annotated with RDF metadata are combined to allow querying, machine learning and inference across different representational domains.
The metadata model considered in this context is based on similar concepts, but the problem addressed is different compared to the problem of dynamic and operational interoperability of SOA services in CPS systems.

\section{Mapping to Natural Language Processing}\label{sec:nlp}
Vector embedding of sub-symbolic relations is a powerful concept often applied in natural language processing (NLP).
Vector embeddings of words, sentences and contexts enable mappings between words in terms of vector operations in an $n$-dimensional space, where typically $n > 100$.
Initially, relatively simple vector space models \cite{turney2010frequency} were used and can represent some important word and document relations \cite{sahlgren2006word}.
Lately, neural network based approaches like Word2Vec \cite{mikolov2013efficient} have shown great performance, and thereby the use of simpler embeddings like one-hot vectors have mostly disappeared.
Word2Vec maps the words (symbols) to a manifold in a vector space, thereby creating model with sub-symbolic representations.
Recent advancements have been achieved using attention models \cite{vaswani2017attention} to create embeddings that produce different mappings for the same word given different contexts \cite{devlin2018bert}.
Sub-symbolic vector embeddings of this type has recently been used for ontology alignment purposes \cite{kolyvakis2018deepalignment}.

Work in the field of machine translation is another important source of examples and guidance.
Translation based on traditional statistics have recently been outdated by neural machine translation (NMT) as the state of the art.
This switch was exemplified by Google, who have been using NMT for their translation service since 2016 \cite{wu2016google}.
An upgrade to the translation system allowed them to translate between unseen language pairs \cite{johnson2017google}, a process they call as zero-shot translation.
The translation system in \cite{wu2016google} uses recurrent neural networks with attention.
More recent translation systems use pure attention models based on the transformer model \cite{vaswani2017attention} to achieve state-of-the-art results \cite{dehghani2018universal}.
All translation systems referenced above are based on word or sub-word input features.
There are also examples of fully character-level convolutional approaches \cite{lee2017fully}.

To achieve good results on NLP tasks, the training protocol is of key importance.
A major recent advancement in NLP is the step to semi-supervised pre-training.
With pre-training, a language model in the form of a neural network is created using a large dataset, which can subsequently be fine-tuned for other problems with little data and computational resources.
One of the latest improvements in semi-supervised pre-training is BERT\footnote{http://jalammar.github.io/illustrated-bert/} \cite{devlin2018bert}, which can be downloaded in pre-trained form.
It is an exciting open problem to adapt these concepts and recent technological advancements to the problem of sub-symbolic ontology alignment and more generally to the M2M translator learning task introduced in Section~II. 

\section{Mapping to Graph Neural Networks}\label{sec:graph}

Graph neural network (GNN) models are relatively new in the machine learning field.
Conventional architectures like recurrent- and convolutional neural networks are based on the assumption that there is repetitive structure in the input.
In contrast, GNNs cannot be based on that assumption because graph data is more irregular in nature, see
\cite{wu2019comprehensive} and \cite{lee2018attention} for an overview of the field.
Several concepts from the image recognition and NLP fields have been adapted to GNNs, like graph convolution \cite{gilmer2017neural}, graph attention \cite{velivckovic2017graph} and graph embeddings \cite{ristoski2016rdf2vec}.
The resulting methods have been successfully used for example to study molecule structures in chemistry and perform traffic route planning \cite{wu2019comprehensive}.

An interesting development in the field of semantic technologies is RDF2Vec \cite{ristoski2016rdf2vec},
which is an extension of the Word2Vec model to graph embeddings.
Much like word embedding, graph embedding is a powerful tool to represent graphs in a metric space, where for example graph clustering and similarity tasks can be addressed.
The Relational Graph Convolutional Network (R-GCN) \cite{schlichtkrull2018modeling} is another interesting recent development for processing of RDF-graphs, which is based on the message-passing network architecture in \cite{gilmer2017neural}.
By allowing for different convolution operators for different kinds of edges the R-GCN represents RDF-data more effectively than if all edges are treated the same.
The R-GCN is validated on entity classification and link prediction tasks.

GNNs are currently actively developed and offer interesting new possibilities to perform graph embeddings and data-driven ontology alignment and mappings between $G^A$, $G^B$, $G^C$ etc needed to address the M2M translator learning task outlined in Section~II.

\section{Discussion: Translator Learning Strategies}

Inspired by the recent developments in NLP it is tempting to adopt an encoder-decoder translation scheme, similar to that in \cite{wu2016google} or \cite{lee2017fully} (see Figure~\ref{fig:learning-e2e}).
These models are typically trained end-to-end (E2E) with pairs of known translations, using a message reconstruction loss on the form
$J(\hat{m}^B, m^B)$.
This is feasible in NLP where large repositories of such translation pairs have been developed.
The M2M translation case is challenging because the repertoire of input representations, ``languages'' and ``dialects'' is diverse and more quickly growing, and data sets contains information about production and operations that typically cannot be publicly shared for the collection of such large data sets.
Thus, we will likely have access to less data and relatively few known translation pairs,
since identifying and tagging these pairs is costly and time consuming, which is challenging for obtaining a scalable on-demand interoperability solution.

Accurate one-to-one word translation is not always possible in NLP.
For example, a round-trip translation of the Swedish word ``Lagom'' to English with Google Translate results in ``Moderate'', followed by ``M{\aa}ttlig'', which is semantically related albeit different (sub-symbolic representations are different in most Swedish native speakers).
This is expected because there is no one-to-one mapping between that Swedish word and a word in the domain of English language.
However, the meaning of the word ``Lagom'' can essentially be explained to an English native speaking person by a longer description, with one or a few follow-up questions needed to validate and further align the interpretation of the concept.
Similarly, some messages in CPS~A might require several messages to be accurately represented in CPS~B, and vice versa.
That is why we define the translators, $T^{AB}$, as an integrated part of the overall SOA of the SoS, such as the aforementioned Arrowhead Framework.
This way it is possible, in principle, that the translator requests or provides additional information needed to proceed with a translation.
For example, although the messages in Listing~\ref{lst:data} refers to the same location, an external information source is needed to identify this relationship, for example as described in \cite{moutinho2018extended}.
It should be noted that NLP translations of this type are currently challenging to learn.

Instead of learning the translators in an E2E fashion, it is possible to use the messages communicated within each system as a starting point.
Even if we cannot expect to have access to large data sets of prealigned A--B message pairs, we do expect high rates of internal messages in each CPS.
Thus, we can optimize the embeddings of the messages in each domain separately and make use of the common environmental degrees of freedom to learn relationships between such embeddings.
For example, vector space embeddings can be learned in the form of latent representations of autoencoders as illustrated in Figure~2,
and in this context methods and concepts that are successfully used for NLP can be reused and further developed.
Translations between the latent spaces of the CPS~A and CPS~B encoders can for example be learned by solving Equation~\ref{eq:loss} using loss/utility functions of the type listed in Table~\ref{lossfunctions}.
Such an autoencoding scheme does not solve the problem of missing translation pairs.
However, with sub-symbolic representations of symbols that are optimized with metadata and data from the physical domain, the problem to learn mappings between symbols is simplified and enables faster convergence and learning with less data.
It also enables clustering and classification of messages, which is useful to improve training and testing protocols.
The use of auxilliary goals have for example helped when solving NLP tasks \cite{devlin2018bert}.

\newlength{\trianglelength}
\setlength{\trianglelength}{0.12\columnwidth}
\newcommand{\encdec}[1]{
	\draw ($#1 + (160:1.06418\trianglelength)$) -- node[right] () {\scriptsize Enc} ($#1+(200:1.06418\trianglelength)$) -- ($#1+(200:0.4\trianglelength)$) -- ($#1 + (160:0.4\trianglelength)$) -- cycle;
	\draw ($#1 + (20:1.06418\trianglelength)$) --  ($#1+(340:1.06418\trianglelength)$) -- ($#1+(340:0.4\trianglelength)$) -- node[right] () {\scriptsize Dec} ($#1 + (20:0.4\trianglelength)$) -- cycle;
	\node[text width = 0.7\trianglelength, align=center] at #1 () {\baselineskip=3pt\scriptsize Latent repr.\par};
}
\tikzset{
	rotate up/.style = {shape border rotate=0},
	rotate down/.style = {shape border rotate=180},
	rotate left/.style = {shape border rotate=270},
	rotate right/.style = {shape border rotate=90},
	>=latex
}

\newcommand{\autoencoderFigure}{
	\begin{subfigure}[t]{0.7\columnwidth}
		\centering
		\begin{tikzpicture}[]
			\coordinate (A center) at (0, 0);
			\coordinate (B center) at (0, -2.3\trianglelength);
			\node[draw, fill=gray!20, trapezium, trapezium angle=70, minimum height=0.5\trianglelength, rotate left, left of = A center] (A Enc) {\scriptsize Enc};
			\node[draw, fill=gray!20, trapezium, trapezium angle=70, minimum height=0.5\trianglelength, rotate right, right of = A center] (A Dec) {\scriptsize Dec};
			\node[draw, fill=gray!20, trapezium, trapezium angle=70, minimum height=0.5\trianglelength, rotate left, left of = B center] (B Enc) {\scriptsize Enc};
			\node[draw, fill=gray!20, trapezium, trapezium angle=70, minimum height=0.5\trianglelength, rotate right, right of = B center] (B Dec) {\scriptsize Dec};
			\node[left = 0.5\trianglelength of A Enc] (ma in) {$\mathrm{m}^A$};
			\node[right = 0.5\trianglelength of A Dec] (ma out) {$\hat{\mathrm{m}}^A$};
			\node[left = 0.5\trianglelength of B Enc] (mb in) {$\mathrm{m}^B$};
			\node[right = 0.5\trianglelength of B Dec] (mb out) {$\hat{\mathrm{m}}^B$};
			\draw[->] (ma in) to (A Enc);
			\draw[->] (A Dec) to (ma out);
			\draw[->] (mb in) to (B Enc);
			\draw[->] (B Dec) to (mb out);
			\draw (A Enc) -- node[pos=0.5, label={[label distance=-0.2\trianglelength] $h^A$}] (A latent) {} (A Dec);
			\draw (B Enc) -- node[pos=0.5, label={[label distance=-0.2\trianglelength] $h^B$}] (B latent) {} (B Dec);
			\node at ($(A latent)!0.5!(B latent)$) (est latent) {$\hat{h}^B$};
			\draw[very thick, ->, bend left] (A Enc.base east) to node[align=center, right] (trans) {\footnotesize$T^{AB}$} (est latent);
			\draw[->, bend right] (est latent) to (B Dec.160);
			\node[align=center] (metadata) at ($(ma in)!0.5!(mb in)$) {$G^A$\\$G^B$};
			\draw[->, bend left] (metadata) to (A Enc.220);
			\draw[->, bend right] (metadata) to (B Enc.140);
		\end{tikzpicture}
		\caption{}
		\label{fig:autoencoders}
	\end{subfigure}
}

\newcommand{\eTeTranslationFigure}{
	\begin{subfigure}[t]{0.6\columnwidth}
		\centering
		\begin{tikzpicture}
			\coordinate (center) at (0, 0);
			\node[draw, fill=gray!20, trapezium, trapezium angle=70, minimum height=0.5\trianglelength, rotate left, left of = center] (Enc) {\scriptsize Enc};
			\node[draw, fill=gray!20, trapezium, trapezium angle=70, minimum height=0.5\trianglelength, rotate right, right of = center] (Dec) {\scriptsize Dec};
			\node[left = 0.5\trianglelength of  Enc] (ma in) {$\mathrm{m}^A$};
			\node[right = 0.5\trianglelength of Dec] (mb out) {$\hat{\mathrm{m}}^B$};
			\draw[->] (Dec) to (mb out);
			\draw[->] (ma in) to (Enc);
			\draw (Enc) -- node[label={[label distance=-0.2\trianglelength] $h^{AB}$}] (latent) {} (Dec);
			\node[fit=(Enc) (Dec), inner sep=10pt, label={$T^{AB}$}] (trans) {};
			\draw[->, very thick] (trans.north west) to (trans.north east);
			\path (ma in) to node[ellipse, below, inner sep=1pt] (phys) {\phantom{\scriptsize$u^A, y^B; u^B, y^A$}} (mb out);
		\end{tikzpicture}
		\vspace{-20pt}
		\caption{}
		\label{fig:learning-e2e}
	\end{subfigure}
}

\newcommand{\eTeTranslationPhysicsFigure}{
	\begin{subfigure}[t]{0.6\columnwidth}
		\centering
		\begin{tikzpicture}
			\coordinate (center) at (0, 0);
			\node[draw, trapezium, trapezium angle=70, minimum height=0.5\trianglelength, rotate left, left of = center] (Enc) {\scriptsize Enc};
			\node[draw, trapezium, trapezium angle=70, minimum height=0.5\trianglelength, rotate right, right of = center] (Dec) {\scriptsize Dec};
			\node[below left = 0.3\trianglelength of  Enc] (ma in) {$\mathrm{m}^A$};
			\node[below right = 0.3\trianglelength of Dec] (mb out) {$\hat{\mathrm{m}}^B$};
			\draw[->] (Dec) -| (mb out);
			\draw[->] (ma in) |- (Enc);
			\draw (Enc) -- node[label={[label distance=-0.2\trianglelength] \scriptsize Latent space}] (latent) {} (Dec);
			\path (ma in) to node[draw, ellipse, below, inner sep=1pt] (phys) {\scriptsize$u^A, y^B; u^B, y^A$} (mb out);
			\draw[->] (latent.center) -- (phys);
			\node[draw, dashed, fit=(Enc) (Dec), label={\footnotesize Translator}, inner sep=5pt] {};
		\end{tikzpicture}
		\caption{End-to-end translation with physical property prediction.}
		\label{fig:learning-e2e-physical}
	\end{subfigure}
}

\newcommand{\handMadeTranslator}{
	\begin{subfigure}[t]{0.6\columnwidth}
		\centering
		\begin{tikzpicture}[>=latex]
			\node[draw, fill=gray!20, rounded corners, text width = 2\trianglelength, align=center] (hand) {\footnotesize Engineered translator};
			\node[left = \trianglelength of hand] (ma in) {};
			\draw[->] (ma in) -- node[above]  {$\mathrm{m}^A$} (hand);
			\node[right = \trianglelength of hand] (ma out) {};
			\draw[->] (hand) -- node[above] {$\hat{\mathrm{m}}^B$} (ma out);
			\node[below = 0.3\trianglelength of hand] (metadata) {$G^A, G^B$};
			\draw[->] (metadata) -- (hand);
		\end{tikzpicture}
		\caption{Grejor}
		\label{fig:grejor}
	\end{subfigure}
}

\begin{figure}
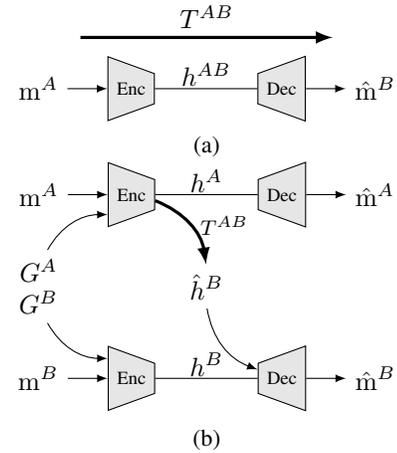

	\centering
	\eTeTranslationFigure
	\vspace{0.5ex}
	\autoencoderFigure
	\caption{Examples of autoencoder translation models. (a) End-to-end translator. (b) Latent representation translator.
	}
	\label{fig:possible_translators}
\end{figure}

A final remark in this discussion concerns the nature of the environment, which up to this point was considered to be reality, so that the mathematical relationships between $u$ and $v$ can be described in terms of physical models.
The translator learning task introduced in Section~II is not limited to natural environments because it only requires that the systems have related degrees of freedom in the environment.
If the transformations from $\{u, v\}$ to $\{u^A, v^A\}$ and $\{u, v\}$ to $\{u^B, v^B\}$ are orthogonal there is little that can be learned using the approach proposed here.
However, in systems of our primary interest  correlations between some $y^A$ and $y^B$ are expected and causal relationships are expected between some $u^A$ and $y^B$, and vice versa.
This is the case also for interconnected simulation models like digital twins and at higher levels and across levels of the (ISA-95) automation pyramid because most systems and services do not function independently of the others.

\section{Concluding remarks}

Industrial IoT and {\ifour} require adaptable solutions to deal with the high heterogeneity of systems and data.
	In this paper, we have presented a mathemathical interoperability model in which we can describe data-, dynamic- and operational interoperability as machine learning tasks.
	Unlike previous works, which focus on interoperability as an engineering task, our model allows engineers to define operational goals which can be used for automatic optimization of translators.
	The model is flexible and can be used with a variety of machine learning tools and methods.

	Using the model, we propose learning strategies based on advances in natural language processing and %
	graph neural networks, allowing for grounded translators.
	Symbol grounding is achieved using sub-symbolic representations learned in a shared environment.
	In this paper we have mostly assumed that the shared environment is physical, but in principle the shared environment could be any environment suitable for fitting sub-symbolic relationships, for example simulations involving digital twins. %
	Using digital twins, translators can be trained and tested virtually, potentially reducing the time-to-deployment and probability of errors.

	While engineered adapters based on ontology alignment and proof engines are explainable, and eventual problems that occur at runtime can be analyzed and solved by the engineers,
	translators generated with machine learning methods can be more challenging to comprehend.
	This is something industry often find undesirable.
	Data availability is also an issue since there are no large public data sets of semantically dissimilar M2M-type messages available as far as we know.

	In future work we aim to address these issues and provide proof of concept of the model and translator learning task using a simulated environment.

	\section{Acknowledgements}
	We thank Magnus Sahlgren for helpful advice in the area of natural language processing and Sergio Martin del Campo for helpful comments on an early version of the manuscript.

\renewcommand{\bibfont}{\footnotesize}
\bibliographystyle{IEEEtranN}
\bibliography{IEEEabrv,new,old,nlp,graph,semantic,system,cps}

\end{document}